\documentclass{article}

\usepackage{times}
\usepackage{graphicx} 
\usepackage{subfigure} 

\usepackage{natbib}

\usepackage{algorithm}
\usepackage{algorithmic}

\usepackage{hyperref}

\usepackage[accepted]{icml2017}

\usepackage{mycustom}

\icmltitlerunning{Bottleneck Conditional Density Estimation}
\usepackage[capitalize]{cleveref}

\begin{document} 

\twocolumn[
\icmltitle{Bottleneck Conditional Density Estimation}

\icmlsetsymbol{equal}{*}
\begin{icmlauthorlist}
\icmlauthor{Rui Shu}{stanford}
\icmlauthor{Hung H. Bui}{adobe}
\icmlauthor{Mohammad Ghavamzadeh}{google} 
\end{icmlauthorlist}

\icmlaffiliation{stanford}{Stanford University}
\icmlaffiliation{adobe}{Adobe Research}
\icmlaffiliation{google}{DeepMind (The work was done when all the authors were with Adobe Research)}

\icmlcorrespondingauthor{Rui Shu}{ruishu@stanford.edu}

\icmlkeywords{machine learning, ICML}

\vskip 0.3in
]

\printAffiliationsAndNotice{}

\begin{abstract} 
We introduce a new framework for training deep generative models for high-dimensional conditional density estimation. The Bottleneck Conditional Density Estimator (BCDE) is a variant of the conditional variational autoencoder (CVAE) that employs layer(s) of stochastic variables as the bottleneck between the input $x$ and target $y$, where both are high-dimensional. Crucially, we propose a new hybrid training method that blends the conditional generative model with a joint generative model. Hybrid blending is the key to effective training of the BCDE, which avoids overfitting and provides a novel mechanism for leveraging unlabeled data. We show that our hybrid training procedure enables models to achieve competitive results in the MNIST quadrant prediction task in the fully-supervised setting, and sets new benchmarks in the semi-supervised regime for MNIST, SVHN, and CelebA.
\end{abstract} 

\section{Introduction}

Conditional density estimation (CDE) refers to the problem of estimating a conditional density $p(y\vert x)$ for the input $x$ and target $y$. In contrast to classification where the target $y$ is simply a discrete class label, $y$ is typically continuous or high-dimensional in CDE. Furthermore, we want to estimate the full conditional density (as opposed to its conditional mean in regression), an important task the conditional distribution has multiple modes. CDE problems in which both $x$ and $y$ are high-dimensional have a wide range of important applications, including video prediction, cross-modality prediction (e.g. image-to-caption), model estimation in model-based reinforcement learning, and so on.   

Classical non-parametric conditional density estimators typically rely on local Euclidean distance in the original input and target spaces \citep{holmes:cde}. This approach quickly becomes ineffective in high-dimensions from both computational and statistical points of view. Recent advances in deep generative models have led to new parametric models for high-dimensional CDE tasks, namely the conditional variational autoencoders (CVAE) \citep{sohn:cvae}. CVAEs have been applied to a variety of problems, such as MNIST quadrant prediction, segmentation \citep{sohn:cvae}, attribute-based image generation \citep{yan:attribute2image}, and machine translation \citep{zhang:vaemt}. 

But CVAEs suffer from two statistical deficiencies. First, they do not learn the distribution of the input $x$. We argue that in the case of high-dimensional input $x$ where there might exist a low-dimensional representation (such as a low-dimensional manifold) of the data, recovering this structure is important, even if the task at hand is to learn the conditional density $p(y|x)$. Otherwise, the model is susceptible to overfitting. Second, for many CDE tasks, the acquisition of labeled points is costly, motivating the need for semi-supervised CDE. A purely conditional model would not be able to utilize any available unlabeled data.\footnote{We define a ``labeled point'' to be a paired $(x, y)$ sample, and an ``unlabeled point'' to be unpaired $x$ or $y$.} We note that while variational methods \citep{kingma:vae,rezende:vae} have been applied to semi-supervised classification (where $y$ is a class label) \citep{kingma:ssl,lars:auxiliary}, semi-supervised CDE (where $y$ is high-dimensional) remains an open problem.

We focus on a set of deep conditional generative models, which we call \emph{bottleneck conditional density estimators} (BCDEs). In BCDEs, the input $x$ influences the target $y$ via layers of bottleneck stochastic variables $z=\set{z_i}$ in the generative path. The BCDE naturally has a joint generative sibling model which we denote the \emph{bottleneck joint density estimator} (BJDE), where the bottleneck $z$ generates $x$ and $y$ independently. Motivated by \citet{bishop:blend}, we propose a hybrid training framework that regularizes the conditionally-trained BCDE parameters toward the jointly-trained BJDE parameters. This is the key feature that enables semi-supervised learning for conditional density estimation in the BCDEs.


Our BCDE hybrid training framework is a novel approach for leveraging unlabeled data for conditional density estimation. Using our BCDE hybrid training framework, we establish new benchmarks for the quadrant prediction task \citep{sohn:cvae} in the semi-supervised regime for MNIST, SVHN, and CelebA. Our experiments show that {\bf 1)} hybrid training is competitive for fully-supervised CDE, {\bf 2)} in semi-supervised CDE, hybrid training helps to avoid overfitting, performs significantly better than conditional training with unlabeled data pre-training, and achieves state-of-the-art results, and {\bf 3)} hybrid training encourages the model to learn better and more robust representations.

\section{Background}
\subsection{Variational Autoencoders}
Variational Autoencoder (VAE) is a deep generative model for density estimation. It consists of a latent variable $z$ with unit Gaussian prior $z\sim \Normal(0,I_k)$, which in turn generates an observable vector $x$. The observation is usually conditionally Gaussian $x \vert z \sim \Normal\big(\mu_\theta(z), \diag(\sigma^2_\theta(z)\big)$, where $\mu$ and $\sigma^2$ are neural networks whose parameters are represented by $\theta$.\footnote{For discrete $x$, one can use a deep network to parameterize a Bernoulli or a discretized logistic distribution.}  VAE can be seen as a non-linear generalization of the probabilistic PCA \citep{tipping:pca}, and thus, can recover non-linear manifolds in the data. However, VAE's flexibility makes posterior inference of the latent variables intractable. This inference issue is addressed via a recognition model $q_\phi(z\vert x)$, which serves as an amortized variational approximation of the intractable posterior $p_\theta(z\vert x)$. Learning in VAE's is done by jointly optimizing the parameters of both the generative and recognition models so as to maximize an objective that resembles an autoencoder regularized reconstruction loss \citep{kingma:vae}, i.e.,
\begin{align}
&\sup_{\theta,\phi} \; \Expect_{q_\phi(z\vert x)} \big[\ln p_\theta(x \vert z)\big] - \KL\big(q_\phi(z\vert x)\;||\;p(z)\big).
\label{eq:vae-obj-ae}
\end{align}
We note that the objective \cref{eq:vae-obj-ae} can be rewritten in the following form that exposes its connection to the variational lower bound of the log-likelihood
\begin{align}
\sup_{\theta} \Big(\ln p_\theta(x) &- \inf_\phi \; \KL\big(q_\phi(z\vert x) \; || \; p_\theta(z\vert x)\big) \Big)\nonumber\\
&=\sup_{\theta,\phi} \; \Expect_{q_\phi(z\vert x)}\brac{ \ln \frac{p_\theta(x, z)}{p_\phi(z \vert x)}}.
\label{eq:vae-obj-posterior-reg}
\end{align}
We make two remarks regarding the minimization of the term $\KL\big(q_\phi(z\vert x) \;||\; p_\theta(z\vert x)\big)$ in Eq. \ref{eq:vae-obj-posterior-reg}. First, when $q(\cdot|\cdot)$ is a conditionally independent Gaussian, this approximation is at best as good as the mean-field approximation that minimizes $\KL\big(q \;||\; p_\theta(z\vert x)\big)$ over all independent Gaussian $q$'s. Second, this term serves as a form of amortized posterior regularization that encourages the posterior $p_\theta(z\vert x)$ to be close to an amortized variational family \citep{dayan:helmholtz,ganchev:posterior,hinton:wakesleep}. In practice, both $\theta$ and $\phi$ are jointly optimized in \cref{eq:vae-obj-ae}, and the reparameterization trick \citep{kingma:vae} is used to transform the expectation over $z\sim {q_{\phi}(z \vert x)}$ into $\epsilon\sim \Normal(0,I_k);\ z=\mu_\phi(x) + \diag\big(\sigma^2_\phi(x)\big)\epsilon$, which leads to an easily obtained stochastic gradient.

\subsection{Conditional VAEs (CVAEs)}
In \citet{sohn:cvae}, the authors introduce the conditional version of variational autoencoders. The conditional generative model is similar to VAE, except that the latent variable $z$ and the observed vector $y$ are both conditioned on the input $x$. The conditional generative path is
\begin{align}
p_\theta(z \mid x) &= \Normal\Big(z \mid \mu_{z,\theta}(x), \text{diag}\big(\sigma^2_{z,\theta}(x)\big)\Big)\\ 
p_\theta(y \mid x, z) &= \Normal\Big(y \mid \mu_{y,\theta}(x, z), \text{diag}\big(\sigma^2_{y,\theta}(x, z)\big)\Big),
\end{align}
and when we use a Bernoulli decoder is
\begin{align}
p_\theta(y \mid x, z) = \Ber\big(y \mid \mu_{y,\theta}(x, z)\big).
\end{align}
Here, $\theta$ denotes the parameters of the neural networks used in the generative path. The CVAE is trained by maximizing a lower bound of the conditional likelihood
\begin{align}
\ln p_\theta(y \vert x) \ge \Expect_{q_\phi(z \vert x, y)}\brac{\ln \frac{p_\theta(z \vert x)p_\theta(y \vert x, z)}{q_\phi(z \vert x, y)}},\label{eq:cvae}
\end{align}
but with a recognition network $q_\phi(z \vert x, y)$, which is typically Gaussian $\Normal\left(z \vert \mu_\phi(x, y), \text{diag}\big(\sigma^2_\phi(x, y)\big)\right)$, and takes both $x$ and $y$ as input. 

\begin{figure*}[!t]
\centering
\begin{tikzpicture}
\node[circ] (zp) {$z$};
\node[circ, xshift=-1.5cm, yshift=-1.5cm] (xp) {$x$};
\node[circ, xshift=+1.5cm, yshift=-1.5cm] (yp) {$y$};
\node[draw, rectangle, rounded corners, minimum width=4.5cm, minimum height=3.5cm, yshift=-0.5cm]    {};
\node[none, yshift=+0.8cm] {BJDE};

\node[circ, xshift=+8cm] (z) {$z$};
\node[circ, xshift=+6.5cm, yshift=-1.5cm] (x) {$x$};
\node[circ, xshift=+9.5cm, yshift=-1.5cm] (y) {$y$};
\node[rrec, xshift=+8cm, yshift=-0.5cm]    {};
\node[none, xshift=+8cm, yshift=+0.8cm] {BCDE};

\node[none, xshift=+4cm] {Regularization};
\node[circ, xshift=+2cm, yshift=-0.3cm, draw=none] (s) {};
\node[circ, xshift=+6cm, yshift=-0.3cm, draw=none] (e) {};
\node[none, xshift=0cm, yshift=-2.3cm] (j) {};
\node[none, xshift=8cm, yshift=-2.3cm] (c) {};
\node[none, xshift=0cm, yshift=-4cm, text width=3cm, align=center] (u) {Unpaired Data\\$\set{x_i} \cup \set{y_i}$};
\node[none, xshift=4cm, yshift=-4cm, text width=3cm, align=center] (p) {Paired Data\\$\set{x_i, y_i}$};

\path
(zp) edge [connect] (xp)
(zp) edge [connect] (yp)
(xp) edge [rdotconnect, bend right=45] (zp)
(yp) edge [ldotconnect, bend left=45] (zp)
(x) edge [connect] (z)
(z) edge [connect] (y)
(x) edge [rdotconnect, bend right=45] (z)
(y) edge [ldotconnect, bend left=45] (z)
(s) edge [bothconnect] (e)
(u) edge [connect] (j)
(p) edge [connect] (j)
(p) edge [connect] (c)
;
\end{tikzpicture}  
\caption{The hybrid training procedure that regularizes BCDE towards BJDE. This regularization enables the BCDE to indirectly leverage unpaired $x$ and $y$ for conditional density estimation.}
\label{fig:bde}
\end{figure*}
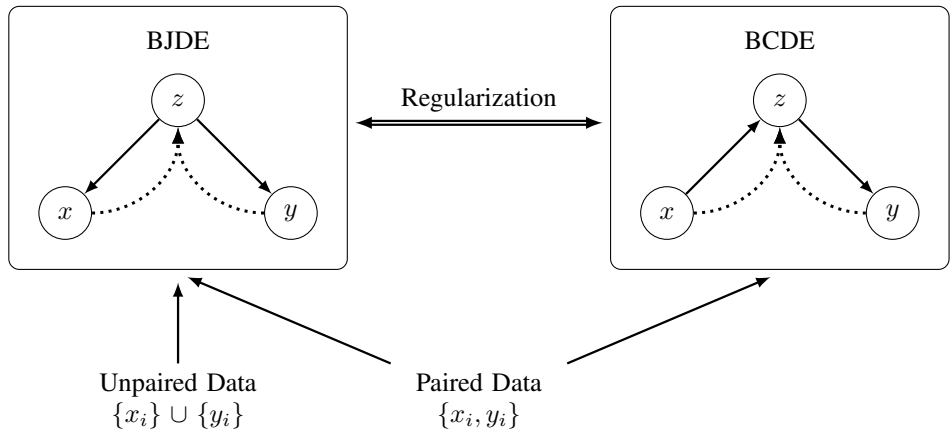
\subsection{Blending Generative and Discriminative}\label{sec:bishop}
It is well-known that a generative model may yield sub-optimal performance when compared to the same model trained discriminatively  \citep{ng:versus}, a phenomenon attributable to the generative model being mis-specified \citep{bishop:blend}. However, generative models can easily handle unlabeled data in semi-supervised setting. This is the main motivation behind blending generative and discriminative models. \citet{bishop:blend} proposed a principled method for hybrid blending by duplicating the parameter of the generative model into a discriminatively trained $\theta$ and a generatively trained $\thetap$, i.e.,
\begin{equation}
p(\X_l, \Y_l, \X_u, \thetap, \theta) = p(\thetap, \theta)p(\X_u \vert \thetap) p(\X_l \vert \thetap) p(\Y_l \vert \X_l, \theta).
\label{eq:bishop-blend}
\end{equation}
%
%
The discriminatively trained parameter $\theta$ is regularized toward the generatively trained parameter $\thetap$ via a prior $p(\thetap,\theta)$ that prefers small $\|\theta-\thetap\|^2$. As a result, in addition to learning from the labeled data $(\X_l,\Y_l)$, the discriminative parameter $\theta$ can be informed by the unlabeled data  $\X_u$ via $\thetap$, enabling a form of semi-supervised discriminatively trained generative model. However, this approach is limited to simple generative models (e.g., naive Bayes and HMMs), where exact inference of $p(y\vert x, \theta)$ is tractable. 

\section{Neural Bottleneck Conditional Density Estimation}
While \citet{sohn:cvae} has successfully applied the CVAE to CDE, CVAE suffers from two limitations. First, the CVAE does not learn the distribution of its input $x$, and thus, is far more susceptible to overfitting. Second, it cannot incorporate unlabeled data. To resolve these limitations, we propose a new approach to high-dimensional CDE that blends the discriminative model that learns the conditional distribution $p(y|x)$, with a generative model that learns the joint distribution $p(x, y)$.

\subsection{Overview}
Figure \ref{fig:bde} provides a high-level overview of our approach that consists of a new architecture and a new training procedure. Our new architecture imposes a bottleneck constraint, resulting a class of conditional density estimators, we call it \emph{bottleneck conditional density estimators} (BCDEs). Unlike CVAE, the BCDE generative path prevents $x$ from directly influencing $y$. Following the conditional training paradigm in \citet{sohn:cvae}, conditional/discriminative training of the BCDE means maximizing the lower bound of a conditional likelihood similar to \eqref{eq:cvae},i.e.,
\begin{align*}
\ln p_\theta(y \vert x) &\ge \C(\theta, \phi; x, y) \\
&=\Expect_{q_\phi(z \vert x, y)}\brac{\ln \frac{p_\theta(z \vert x)p_\theta(y \vert z)}{q_\phi(z \vert x, y)}}.
\end{align*}
When trained over a dataset of paired $(\X, \Y)$ samples, the overall conditional training objective is
\begin{align}
\C(\theta, \phi; \X, \Y) &= \sum_{x, y \in \X, \Y} \C(\theta, \phi; x, y). \label{eq:cond}
\end{align}
However, this approach suffers from the same limitations as CVAE and imposes a bottleneck that limits the flexibility of the generative model. Instead, we propose a {\em hybrid} training framework that takes advantage of the bottleneck architecture to avoid overfitting and supports semi-supervision.

One component in our hybrid training procedure tackles the problem of estimating the \emph{joint} density $p(x,y)$. To do this, we use the joint counterpart of the BCDE: the bottleneck joint density estimator (BJDE). Unlike conditional models, the BJDE allows us to incorporate unpaired $x$ and $y$ data during training. Thus, the BJDE can be trained in a semi-supervised fashion. We will also show that the BJDE is well-suited to \emph{factored inference} (see \cref{sec:factor}), i.e., a factorization procedure that makes the parameter space of the recognition model more compact.

The BJDE also serves as a way to regularize the BCDE, where the regularization constraint can be viewed as soft-tying between the parameters of these two models' generative and recognition networks. Via this regularization, BCDE benefits from unpaired $x$ and $y$ for conditional density estimation.

\begin{figure*}
\centering
\subfigure[Joint: $(x)$]{
\begin{tikzpicture}
\node[none, xshift=1cm] () {};
\node[none, xshift=-1cm] () {};
\node[circ] (z) {$z$};
\node[circ, yshift=-2cm] (x) {$x$};
\path
(z) edge [connect] (x)
(x) edge [rdotconnect, bend right=45] (z)
;
\end{tikzpicture}
\label{fig:vaex}
}
\subfigure[Joint: $(y)$]{
\begin{tikzpicture}
\node[circ] (z) {$z$};
\node[circ, xshift=+2cm] (y) {$y$};
\node[circ, draw=none, yshift=-2cm] (x) {};
\path
(z) edge [connect] (y)
(y) edge [ldotconnect, bend left=45] (z)
;
\end{tikzpicture}
\label{fig:vaey}
}
\subfigure[Joint: $(x, y)$]{
\begin{tikzpicture}
\node[none, xshift=-.7cm] () {};
\node[circ] (z) {$z$};
\node[circ, yshift=-2cm] (x) {$x$};
\node[circ, xshift=+2cm] (y) {$y$};
\path
(z) edge [connect] (x)
(z) edge [connect] (y)
(x) edge [rdotconnect, bend right=45] (z)
(y) edge [ldotconnect, bend left=45] (z)
;
\end{tikzpicture}
\label{fig:vaexy}
}
\subfigure[Conditional: $(x, y)$]{
\begin{tikzpicture}
\node[none, xshift=3.5cm] () {};
\node[none, xshift=-1cm] () {};
\node[circ] (z) {$z$};
\node[circ, yshift=-2cm] (x) {$x$};
\node[circ, xshift=+2cm] (y) {$y$};
\path
(x) edge [connect] (z)
(z) edge [connect] (y)
(x) edge [rdotconnect, bend right=45] (z)
(y) edge [ldotconnect, bend left=45] (z)
;
\end{tikzpicture}
\label{fig:cvae}
}
\caption{The joint and conditional components of the BCDE. Dotted lines represent recognition models. The conditional model parameters are regularized toward the joint model's. The natural pairing of the conditional and joint parameters is described in \cref{table:shared}.}\label{fig:models}
\end{figure*}
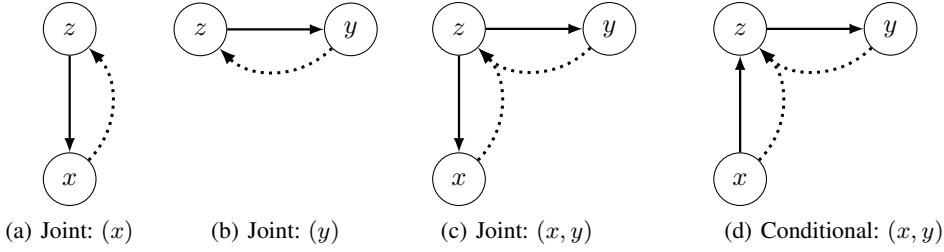
\begin{table*}[!t]
\centering
\begin{tabular}{llccccc}
\toprule
\multirow{2}{*}{Standard}
& BJDE: & 
$q_\phip(z \vert x, y)$ &
$q_\phip(z \vert y)$ & 
$q_\phip(z \vert x)$ & 
$p_\thetap(y \vert z)$ &
$p_\thetap(x \vert z)$  
\\
\cmidrule{2-7} 
& BCDE: &
$q_\phi(z \vert x, y)$ &
$-$ & 
$p_\theta(z \vert x)$ & 
$p_\theta(y \vert z)$ &
$-$ 
\\
\midrule
\multirow{2}{*}{Factored}
& BJDE: & 
$-$ &
$\hat{\ell}_\phip(z ; y)$ & 
$q_\phip(z \vert x)$ & 
$p_\thetap(y \vert z)$ &
$p_\thetap(x \vert z)$  
\\
\cmidrule{2-7} 
& BCDE: &
$-$ &
$\hat{\ell}_\phip(z ; y)$ & 
$p_\theta(z \vert x)$ & 
$p_\theta(y \vert z)$ &
$-$ 
\\
\midrule
\end{tabular}
\caption{Soft parameter tying between the BJDE and BCDE. For each network within the BCDE, there is a corresponding network within the BJDE. We show the correspondence among the networks with and without the application of factored inference. We regularize all the BCDE networks to their corresponding BJDE network parameters.} 
\label{table:shared}
\end{table*}

\subsection{Bottleneck Joint Density Estimation}
In the BJDE, we wish to learn the joint distribution of $x$ and $y$. The bottleneck is introduced in the generative path via the bottleneck variable $z$, which points to $x$ and $y$ (see \cref{fig:vaex,fig:vaey,fig:vaexy}). Thus, the variational lower bound of the joint likelihood is
\begin{align}
\ln p_\thetap(x, y) &\ge \J_{xy}(\thetap, \phip; x, y) \nonumber \\
&=\Expect_{q_\phip(z \vert x, y)} \brac{\ln \frac{p(z)p_\thetap(x \vert z) p_\thetap(y \vert z)}{q_\phip(z \vert x, y)}}.
\end{align}
We use $\{\thetap, \phip\}$ to indicate the parameters of the BJDE networks and reserve $\set{\theta, \phi}$ for the BCDE parameters. For samples in which $x$ or $y$ is unobserved, we will need to compute the variational lower bound for the marginal likelihoods. Here, the {\em bottleneck} plays a critical role. If $x$ were to directly influence $y$ in a non-trivial manner, any attempt to incorporate unlabeled $y$ would require the recognition model to infer the unobserved $x$ from the observed $y$\textemdash a conditional density estimation problem which might be as hard as our original task. In the bottleneck architecture, the conditional independence of $x$ and $y$ given $z$ implies that only the low-dimensional bottleneck needs to be marginalized. Thus, the usual variational lower bounds for the marginal likelihoods yield
\begin{align}
\ln p_\thetap(x) &\ge 
\J_x(\thetap, \phip; x) = 
\Expect_{q_\phip(z \vert x)} \brac{\ln \frac{p(z)p_\thetap(x \vert z)}{q_\phip(z \vert x)}},\nonumber \\
\ln p_\thetap(y) &\ge 
\J_y(\thetap, \phip; y) = 
\Expect_{q_\phip(z \vert y)} \brac{\ln \frac{p(z)p_\thetap(y \vert z)}{q_\phip(z \vert y)}}. \nonumber
\end{align}
Since $z$ takes on the task of reconstructing both $x$ and $y$, the BJDE is sensitive to the distributions of $x$ and $y$ and learns a joint manifold over the two data sources. Thus, the BJDE provides the following benefits: \textbf{1}) learning the distribution of $x$ makes the inference of $z$ given $x$ robust to perturbations in the inputs, \textbf{2}) $z$ becomes a joint-embedding of $x$ and $y$, \textbf{3}) the model can leverage unlabeled data. Following the convention in \cref{eq:cond}, the joint training objectives is
\begin{align}
&\J(\thetap, \phip; \X_u, \Y_u, \X_l, \Y_l) =  \label{eq:joint}\\
&\phantom{\hspace{0.5cm}} \J_x(\thetap, \phip; \X_u) + \J_y(\thetap, \phip; \Y_u) + \J_{xy}(\thetap, \phip; \X_l, \Y_l), \nonumber
\end{align}
where $(\X_l, \Y_l)$ is a dataset of paired $(x, y)$ samples, and $\X_u$ and $\Y_u$ are datasets of unpaired samples.  

\subsection{Blending Joint and Conditional Deep Models}
Because of potential model mis-specifications, the BJDE is not expected to yield good performance if applied to the conditional task. Thus, we aim to blend the BJDE and BCDE models in the spirit of \citet{bishop:blend}. However, we note that \eqref{eq:bishop-blend} is not directly applicable since the BCDE and BJDE are two different models, and not two different views (discriminative and generative) of the same model. Therefore, it is not immediately clear how to tie the BCDE and BJDE parameters together. Further, these models involve conditional probabilities parameterized by deep networks and have no closed form for inference. 

Any natural prior for the BCDE parameter $\theta$ and the BJDE parameter $\thetap$ should encourage $p_{\text{BCDE}}(y\vert x,\theta)$ to be close to $p_{\text{BJDE}}(y\vert x, \thetap)$. In the presence of the latent variable $z$, it is then natural to encourage $p(z \vert x, \theta)$ to be close to $p(z \vert x, \thetap)$ and $p(y \vert z, \theta)$ to be close to $p(y \vert z, \thetap)$. However, enforcing the former condition is intractable as we do not have a closed form for $p_{\text{BJDE}}(z \vert x, \thetap)$. Fortunately, an approximation of $p_{\text{BJDE}}(z \vert x, \thetap)$ is provided by the recognition model $q(z \vert x, \phip)$. Thus, we propose to softly tie together the parameters of networks defining $p(z\vert x, \theta)$ and $q(z\vert x, \phip)$. This strategy effectively leads to a joint prior over the model network parameters, as well as the recognition network parameters $p(\phip, \thetap, \phi, \theta)$.





As a result, we arrive at the following hybrid blending of deep stochastic models and its variational lower bound
\begin{align}
&\ln p(\X_l, \Y_l, \X_u, \Y_u, \thetap, \phip, \theta, \phi) \ge \ln p(\thetap, \phip, \theta, \phi) ~+ \nonumber \\
&\phantom{\hspace{1cm}} \J_x(\thetap, \phip; \X_u) + \J_y(\thetap, \phip; \Y_u) ~+ \nonumber \\
&\phantom{\hspace{1cm}} \J_x(\thetap, \phip; \X_l) + \C(\theta, \phi; \X_l, \Y_l).
\label{eq:hybrid}
\end{align}
We interpret $\ln p(\thetap, \phip, \theta, \phi)$ as a $\ell_2$-regularization term that softly ties the joint parameters $(\thetap, \phip)$ and conditional parameters $(\theta, \phi)$ in an appropriate way. For the BCDE and BJDE, there is a natural one-to-one mapping from the conditional parameters to a subset of the joint parameters. For the joint model described in \cref{fig:vaexy} and conditional model in \cref{fig:cvae}, the parameter pairings are provided in \cref{table:shared}. Formally, we define $\gamma = \{\theta, \phi\}$ and use the index $\gamma_{a\vert b}$ to denote the parameter of the neural network on the Bayesian network link $b\rightarrow a$ in the BCDE. For example $\gamma_{z|x}=\theta_{z|x}$, $\gamma_{z|x,y}=\phi_{z|x,y}$. Similarly, let $\gammap = \{\thetap, \phip\}$. In the BJDE, the same notation yields $\gammap_{z|x}=\phip_{z|x}$. The hybrid blending regularization term can be written as 
\begin{align}
\ln p(\theta, \phi, \thetap, \phip) = -\frac{\lambda}{2} \sum_{i\in I} \| \gamma_i - \gammap_i \|_2^2 + \text{const},
\end{align}
where $I$ denotes the set of common indices of the joint and conditional parameters. When the index is $z\vert x$, it effectively means that $p(z \vert x, \theta)$ is softly tied to $q(z \vert x, \thetap)$, i.e.,
\begin{align*}
\| \gamma_{z \vert x} - \gammap_{z \vert x} \|_2^2 = \| \theta_{z \vert x} - \phip_{z \vert x} \|_2^2\;.
\end{align*}
Setting $\lambda=0$ unties the BCDE from the BJDE, and effectively yields to a conditionally trained BCDE, while letting $\lambda \rightarrow \infty$ forces the corresponding parameters of the BCDE and BJDE to be identical. 

Interestingly, \cref{eq:hybrid} does not contain the term $\J_{xy}$. Since explicit training of $\J_{xy}$ may lead to learning a better joint embedding in the space of $z$, we note the following generalization of \cref{eq:hybrid} that trades off the contribution between $\J_{xy}$ and $\brac{\J_{x} + \C}$,
\begin{align}
&\ln p(\X_l, \Y_l, \X_u, \Y_u, \thetap, \phip, \theta, \phi) \nonumber \\
&\phantom{\hspace{0.3cm}}\ge \mathcal{H}(\thetap, \phip, \theta, \phi; \X_l, \Y_l, \X_u, \Y_u) \nonumber \\
&\phantom{\hspace{0.3cm}}=\ln p(\thetap, \phip, \theta, \phi) ~+ \nonumber \\
&\phantom{\hspace{1cm}} \J_x(\thetap, \phip; \X_u) + \J_y(\thetap, \phip; \Y_u) ~+ \nonumber \\
&\phantom{\hspace{1cm}} \alpha\cdot \J_{xy}(\thetap, \phip; \X_l, \Y_l) ~+ \nonumber \\
&\phantom{\hspace{1cm}} (1 - \alpha)\cdot\brac{\J_x(\thetap, \phip; \X_l) + \C(\theta, \phi; \X_l, \Y_l)}.
\label{eq:final-hybrid}
\end{align}
Intuitively, the equation computes the lower bound of $p(\X_l, \Y_l)$, either using the joint parameters $\thetap, \phip$ or factorizes $p(\X_l, \Y_l)$ into $p(\X_l) p(\Y_l \mid \X_l)$ before computing the lower bound of $p(\Y_l \mid \X_l)$ with the conditional parameters. A proof that the lower bound holds for any $0\le\alpha\le 1$ is provided in \cref{sec:derivation}. For simplicity, we set $\alpha=0.5$ and do not tune $\alpha$ in our experiments.

\subsection{Factored Inference}\label{sec:factor}
The inference network $q_\phi(z \vert x, y)$ is usually parameterized as a single neural network that takes both $x$ and $y$ as input. Using the precision-weighted merging scheme proposed by \citet{sonderby:lvae}, we also consider an alternative parameterization of $q_\phi(z \vert x, y)$ that takes a weighted-average of the Gaussian distribution $q_\phi(z \vert x)$ and a Gaussian likelihood term $\hat{\ell}(z; y)$ (see \cref{sec:factored}). Doing so offers a more compact recognition model and more sharing parameters between the BCDE and BJDE (e.g., see the bottom two rows in \cref{table:shared}), but at the cost of lower flexibility for the variational family $q_\phi(z \vert x, y)$.

\begin{table*}[!h]
\small
\centering
\begin{tabular}{lcccc}
\toprule
Models & $n_l=50000$ & $n_l=25000$  & $n_l=10000$ & $n_l=5000$ \\
\midrule
CVAE \cite{sohn:cvae}
& $63.91$ & - & - & - \\
BCDE (conditional)
& $62.45 \pm 0.02$ & $64.50 \pm 0.03$ & $68.23 \pm 0.05$ & $71.66 \pm 0.06$ \\
BCDE (na\"ive pre-train)
& $\bf62.00 \pm 0.02$ & $63.27 \pm 0.04$ & $65.14 \pm 0.05$ & $67.13 \pm 0.04$ \\
BCDE (hybrid)
& $62.16\pm 0.03$ & $\bf 62.90 \pm 0.02$ & $\bf 64.08 \pm 0.03$ & $65.10 \pm 0.03$ \\
BCDE (hybrid + factored)
& $62.81 \pm 0.05$ & $63.47 \pm 0.02$ & $64.16 \pm 0.02$ & $\bf 64.64 \pm 0.05$ \\ \midrule
\end{tabular}
\vspace{-8pt}
\caption{MNIST quadrant prediction task: 1-quadrant. We report the test set loss (IW=$100$) and standard error.} 
\label{table:q1}
\vspace{10pt}
\begin{tabular}{lcccc}
\toprule
Models & $n_l=50000$ & $n_l=25000$  & $n_l=10000$ & $n_l=5000$ \\
\midrule
CVAE \cite{sohn:cvae}
& $44.73$ & - & - & - \\
BCDE (conditional)
& $43.91 \pm 0.01$ & $45.49 \pm 0.03$ & $48.16\pm 0.02$ & $50.83\pm 0.04$ \\
BCDE (na\"ive pre-train)
& $\bf43.53 \pm 0.02$ & $44.42 \pm 0.04$ & $45.81\pm 0.01$ & $47.49\pm 0.06$ \\
BCDE (hybrid)
& $\bf43.56 \pm 0.02$ & $\bf44.10 \pm 0.02$ & $45.23\pm 0.02$ & $46.39\pm 0.03$ \\
BCDE (hybrid + factored)
& $44.07 \pm 0.02$ & $44.41 \pm 0.02$ & $\bf45.02\pm 0.04$ & $\bf45.86\pm 0.06$ \\
\midrule
\end{tabular}
\vspace{-8pt}
\caption{MNIST quadrant prediction task: 2-quadrant.}
\label{table:q2}
\vspace{10pt}
\begin{tabular}{lcccc}
\toprule
Models & $n_l=50000$ & $n_l=25000$  & $n_l=10000$ & $n_l=5000$ \\
\midrule
CVAE \cite{sohn:cvae}
& $20.95$ & - & - & - \\
BCDE (conditional)
& $20.64 \pm 0.01$ & $21.27 \pm 0.01$ & $22.44 \pm 0.03$ & $23.72 \pm 0.04$ \\
BCDE (na\"ive pre-train)
& $20.37 \pm 0.01$ & $20.87 \pm 0.02$ & $21.65 \pm 0.02$ & $22.32 \pm 0.05$ \\
BCDE (hybrid)
& $\bf20.31 \pm 0.01$ & $20.69 \pm 0.02$ & $21.36 \pm 0.02$ & $22.27 \pm 0.02$ \\
BCDE (hybrid + factored)
& $20.43 \pm 0.01$ & $\bf20.56 \pm 0.01$ & $\bf21.16 \pm 0.01$ & $\bf21.81 \pm 0.03$ \\ \midrule 
\end{tabular}
\vspace{-8pt}
\caption{MNIST quadrant prediction task: 3-quadrant.}
\label{table:q3}
\vspace{10pt}
\begin{minipage}{0.48\textwidth}
\begin{tabular}{lcc}
\toprule
Models & $n_l=10000$ & $n_l=5000$ \\
\midrule
BCDE (conditional)
& $4657 \pm 48$ & $4845 \pm 33$ \\
BCDE (na\"ive pre-train)
& $4547 \pm 23$ & $4627 \pm 13$ \\
BCDE (hybrid)
& $\bf4213 \pm 21$ & $\bf4392 \pm 13$ \\
BCDE (hybrid + factored)
& $4700 \pm 146$ & $5030 \pm 165$  \\
\midrule
\end{tabular}
\vspace{-8pt}
\caption{SVHN prediction task: Top-Down.}
\label{table:svhn}
\end{minipage}
\begin{minipage}{0.48\textwidth}
\begin{tabular}{lcc}
\toprule
Models & $n_l=20000$ & $n_l=10000$ \\
\midrule
BCDE (conditional)
& $5805 \pm 2$ & $5817 \pm 3$ \\
BCDE (na\"ive pre-train)
& $5784.8 \pm 0.5$ & $5793 \pm 1$ \\
BCDE (hybrid)
& $5778.6 \pm 0.4$ & $5781.3 \pm 0.5$ \\
BCDE (hybrid + factored)
& $\bf5776.1 \pm 0.3$ & $\bf5780.3 \pm 0.6$  \\
\midrule
\end{tabular}
\vspace{-8pt}
\caption{CelebA prediction task: Top-Down.}
\label{table:celeba}
\end{minipage}
\end{table*}
\section{Experiments}
We evaluated the performance of our hybrid training procedure on the permutation-invariant quadrant prediction task \citep{sohn:multimodal,sohn:cvae} for MNIST, SVHN, and CelebA. The quadrant prediction task is a conditional density estimation problem where an image data set is partially occluded. The model is given the observed region and is evaluated by its perplexity on the occluded region. The quadrant prediction task consists of four sub-tasks depending on the degree of partial observability. 1-quadrant prediction: the bottom left quadrant is observed. 2-quadrant prediction: the left half is observed. 3-quadrant prediction: the bottom right quadrant is \emph{not} observed. Top-down prediction: the top half is observed. 

In the fully-supervised case, the original MNIST training set $\set{x_i'}_{i=1}^{50000}$ is converted into our CDE training set $\set{\X_l, \Y_l} = \set{x_i, y_i}_{i=1}^{50000}$ by splitting each image into its observed $x$ and unobserved $y$ regions according to the quadrant prediction task. Note that the training set does not contain the original class label information. In the $n_l$-label semi-supervised case, we randomly sub-sampled $n_l$ pairs to create our labeled training set $\set{x_i, y_i}_{i=1}^{n_l}$. The remaining $n_u$ paired samples are decoupled and put into our unlabeled training sets $\X_u = \set{x_i}_{i=1}^{n_u}, \Y_u = \set{y_i}_{i=1}^{n_u}$. Test performance is the conditional density estimation performance on the entire test set, which is also split into input $x$ and target $y$ according to the quadrant prediction task. Analogous procedure is used for SVHN and CelebA. 

For comparison against \citet{sohn:cvae}, we evaluate the performance of our models on the MNIST 1-quadrant, 2-quadrant, and 3-quadrant prediction tasks. The MNIST digits are statically-binarized by sampling from the Bernoulli distribution according to their pixel values \citep{salakhutdinov:dbn}. We use a sigmoid layer to learn the parameter of the Bernoulli observation model.

We provide the performance on the top-down prediction task for SVHN and CelebA. We used a discretized logistic observation model \citet{kingma:iaf} to model the pixel values for SVHN and a Gaussian observation model with fixed variance for CelebA. For numerical stability, we rely on the implementation of the discretized logistic distribution described in \citet{salimans:pixel}.

In all cases, we extracted a validation set of $10000$ samples for hyperparameter tuning. While our training objective uses a single (IW=$1$) importance-weighted sample \cite{burda:iwae}, we measure performance using IW=$100$ to get a tighter bound on the test log-likelihood \citep{sohn:cvae}. We run replicates of all experiments and report the mean performance with standard errors. For a more expressive variational family \citep{ranganath:hvm}, we use two stochastic layers in the BCDE and perform inference via top-down inference \cite{sonderby:lvae}. We use multi-layered perceptrons (MLPs) for MNIST and SVHN, and convolutional neural networks (CNNs) for CelebA. All neural networks are batch-normalized \citep{ioffe:batchnorm} and updated with \emph{Adam} \citep{kingma:adam}. The number of training epochs is determined based on the validation set.  The dimensionality of each stochastic layer is $50$, $100$, and $300$ for MNIST, CelebA, and SVHN respectively. All models were implemented in Python\footnote{\href{https://github.com/ruishu/bcde}{github.com/ruishu/bcde}} using Tensorflow \citep{tensorflow}.

\subsection{Conditional Log-Likelihood Performance}
\Cref{table:q1,table:q2,table:q3,table:svhn,table:celeba} show the performance comparisons between the CVAE and the BCDE. For baselines, we use the CVAE, the BCDE trained with the conditional objective, and the BCDE initialized via pre-training $\J_x(\cdot)$ and $\J_y(\cdot)$ using the available $x$ and $y$ data separately (and then trained conditionally). Against these baselines, we measure the performance of the BCDE (with and without factored inference) trained with the hybrid objective $\mathcal{H}(\cdot)$. We tuned the regularization hyperparameter $\lambda=\set{10^{-3}, 10^{-2}, \ldots, 10^{3}}$ on the MNIST 2-quadrant semi-supervised tasks and settled on using $\lambda = 10^{-2}$ for all tasks.

\textbf{Fully-supervised regime}. By comparing in the fully-supervised regime for MNIST (\cref{table:q1,table:q2,table:q3}, $n_l = 50000$), we show that the hybrid BCDE achieves competitive performance against the pretrained BCDE and out-performs previously reported results for CVAE \cite{sohn:cvae}. 

\textbf{Semi-supervised regime}. As the labeled training size $n_l$ reduces, the benefit of having the hybrid training procedure becomes more apparent. The BCDEs trained with the hybrid objective function tend to significantly improve upon its conditionally-trained counterparts. 

On MNIST, hybrid training of the factored BCDE achieves the best performance. Both hybrid models achieve over a 1-nat difference than the pre-trained baseline in some cases\textemdash a significant difference for binarized MNIST \cite{wu:decoder}. Conditional BCDE performs very poorly in the semi-supervised tasks due to overfitting. 

On CelebA, hybrid training of the factored BCDE also achieves the best performance. Both hybrid models significantly out-perform the conditional baselines and yield better visual predictions than conditional BCDE (see \cref{sec:visualization}). The hybrid models also outperform pre-trained BCDE with only half the amount of labeled data. 

On SVHN, the hybrid BCDE with standard inference model significantly out-performs the conditional baselines. However, the use of factored inference results in much poorer performance. Since the decoder is a discretized logistic distribution with learnable scale, it is possible that the factored inference model is not expressive enough to model the posterior distribution. 

\begin{figure}[!h]
\centering
\subfigure[Conditional]{\includegraphics[width=0.23\textwidth]{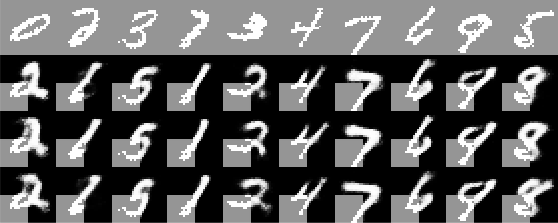}}
\subfigure[Hybrid]{\includegraphics[width=0.23\textwidth]{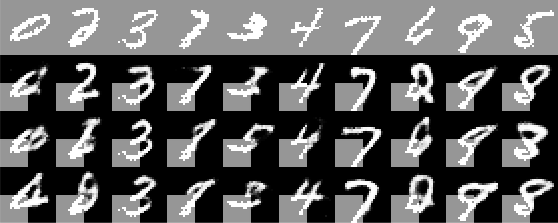}}
\caption{Comparison of conditional image generation for the conditional versus hybrid BCDE on the semi-supervised 1-quadrant task. Row 1 shows the original images.  Rows 2-4 show three attempts by each model to sample $y$ according to $x$ (the bottom-left quadrant, indicated in gray). Hybrid training yields a higher-entropy model that has lower perplexity.}
\label{fig:entropy}
\end{figure}

\textbf{Model entropy.} In \Cref{fig:entropy}, we sample from $p_\theta(y \vert x)$ for the conditional BCDE and the hybrid BCDE. We show that the conditionally-trained BCDE achieves poorer performance because it learns a lower-entropy model. In contrast, hybrid training learns a lower perplexity model, resulting in a high-entropy conditional image generator that spreads the conditional probability mass over the target output space \citep{theis:note}.

\subsection{Conditional Training Overfits}
\begin{figure}[!h]
\centering
\includegraphics[width=0.48\textwidth]{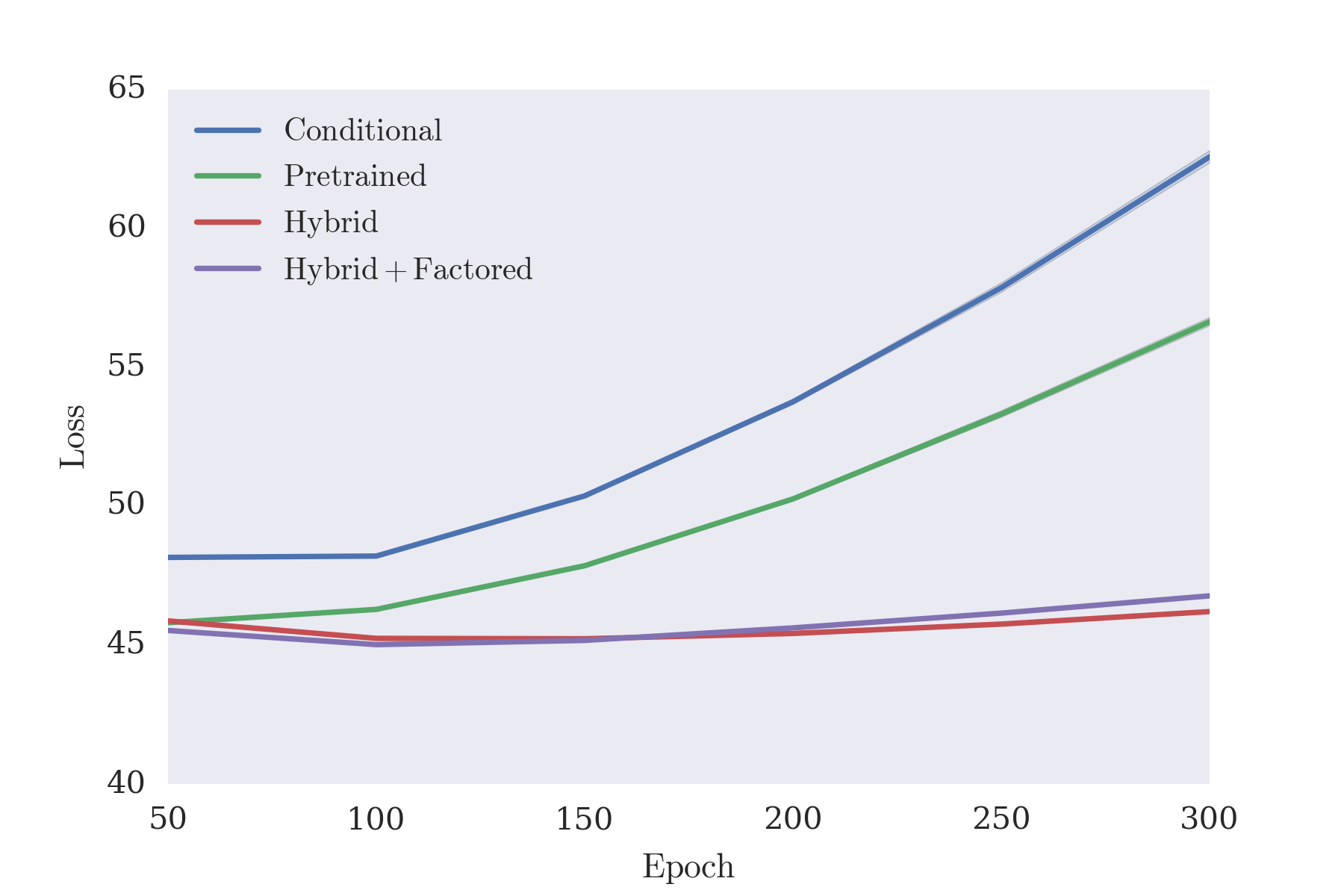}
\caption{Comparison of the BCDE variants on the 2-quadrant MNIST prediction task with $n_l = 10000$ labeled points. In contrast to conditional training, hybrid training is less susceptible to overfitting.}
\label{fig:overfit}
\end{figure}

To demonstrate the hybrid training's regularization behavior, we show the test set performance during training (\cref{fig:overfit}) on the 2-quadrant MNIST task ($n_l = 10000$). Even with pre-trained initialization of parameters, models that were trained conditionally quickly overfit, resulting in poor test set performance. In contrast, hybrid training regularizes the conditional model toward the joint model, which is much more resilient against overfitting.  

\subsection{Robustness of Representation}
Since hybrid training encourages the BCDE to consider the distribution of $x$, we can demonstrate that models trained in a hybrid manner are robust against structured perturbations of the data set. To show this, we experimented with two variants of the MNIST quadrant task called the \emph{shift-sensitive} and \emph{shift-invariant} top-bottom prediction tasks. In these experiments, we set $\lambda = 0.1$.

\subsubsection{Shift-Sensitive Estimation}
In the shift-sensitive task, the objective is to learn to predict the bottom half of the MNIST digit ($y$) when given the top half ($x$). However, we introduce structural perturbation to the top and bottom halves of the image in our training, validation, and test sets by randomly shifting each pair $(x, y)$ horizontally by the same number of pixels (shift varies between $\set{-4, -3, \ldots, 3, 4}$). We then train the BCDE using either the conditional or hybrid objective in the fully-supervised regime. Note that compared to the original top-down prediction task, the perplexity of the conditional task remains the same after the perturbation is applied. 

\begin{table}[!h]
\centering
\small
\begin{tabular}{lccc}
\toprule
Models & No Shift & Shift  & $\Delta$  \\
\midrule
Conditional
& $41.59 \pm 0.02$ & $44.02 \pm 0.03$ & $2.43$ \\
Hybrid
& $41.33 \pm 0.01$ & $43.51 \pm 0.01$ & $2.17$ \\
Hybrid + Factored
& $41.20 \pm 0.02$ & $43.19 \pm 0.02$ & $1.99$ \\
\midrule
\end{tabular}
\caption{Shift-sensitive top-bottom MNIST prediction. Performance with and without structural corruption reported, along with the performance difference. Hybrid training is robust against structural perturbation of $(x, y)$.}
\label{table:sensitive}
\end{table}

\Cref{table:sensitive} shows that hybrid training consistently achieves better performance than conditional training. Furthermore, the hybridly trained models were less affected by the introduction of the perturbation, demonstrating a higher degree of robustness. Because of its more compact recognition model, hybrid + factored is less vulnerable to overfitting, resulting in a smaller performance gap between performance on the shifted and original data.

\subsubsection{Shift-Invariant Estimation}
The shift-invariant task is similar to the shift-sensitive top-bottom task, but with one key difference: we \emph{only} introduce structural noise to the top half of the image in our training, validation, and test sets. The goal is thus to learn that the prediction of $y$ (which is always centered) is invariant to the shifted position of $x$. 

\begin{table}[!h]
\centering
\small
\begin{tabular}{lccc}
\toprule
Models & No Shift & Shift  & $\Delta$  \\
\midrule
Conditional
& $41.59 \pm 0.02$ & $42.99 \pm 0.04$ & $1.40$ \\
Hybrid
& $41.33 \pm 0.01$ & $42.53 \pm 0.02$ & $1.20$ \\
Hybrid + Factored
& $41.20 \pm 0.02$ & $42.20 \pm 0.02$ & $1.00$ \\
\midrule
\end{tabular}
\caption{Shift-invariant top-bottom MNIST prediction. Performance with and without structural corruption reported, along with the performance difference. Hybrid training is robust against structural corruption of $x$.}
\label{table:invariant}
\end{table}
\begin{figure}[!h]
\centering
\includegraphics[width=0.48\textwidth]{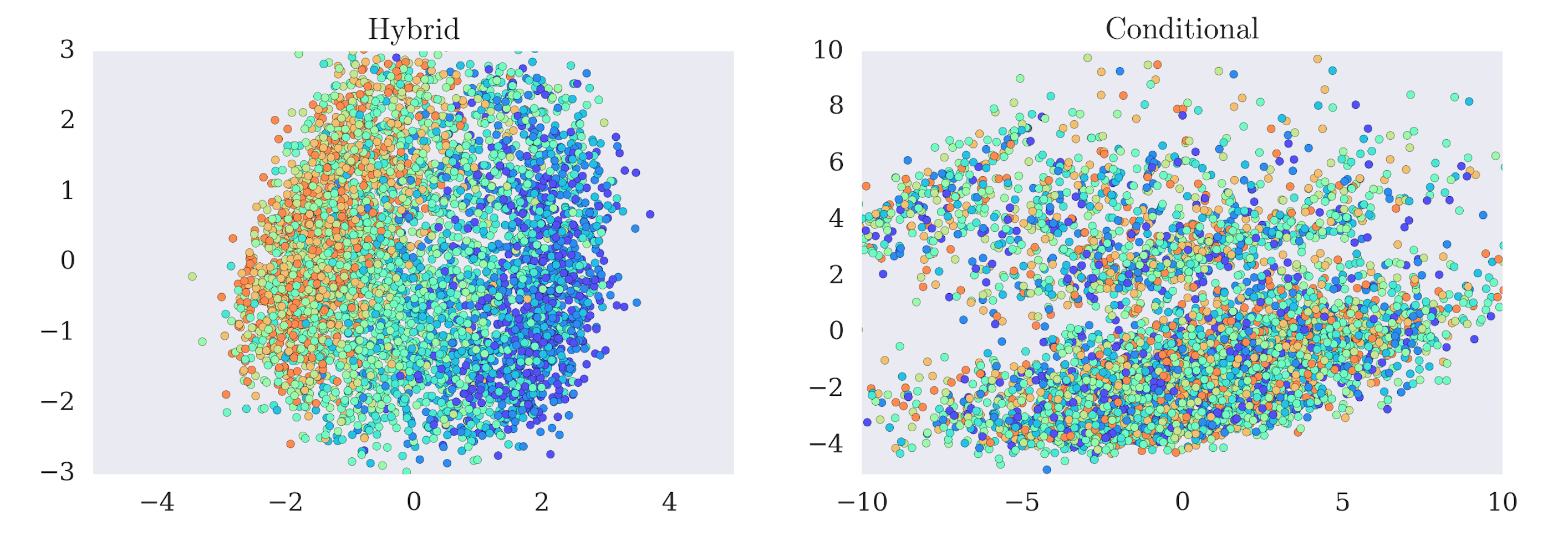}
\caption{Visualization of the latent space of hybrid and conditionally-trained BCDEs. PCA plots of the latent space subregion for all $x$'s whose class label $=2$ are shown. Fill color indicates the degree of shift: blue $=-4$, orange $=+4$.}
\label{fig:invariant}
\end{figure}

\Cref{table:invariant} shows similar behavior to \Cref{table:sensitive}. Hybrid training continues to achieve better performance than conditional models and suffer a much smaller performance gap when structural corruption in $x$ is introduced. 

In \cref{fig:invariant}, we show the PCA projections of the latent space sub-region populated by digits $2$ and color-coded all points based on the degree of shift. We observe that hybrid training versus conditional training of the BCDE result in very different learned representations in the stochastic layer. Because of regularization toward the joint model, the hybrid BCDE's latent representation retrains information about $x$ and learns to untangle shift from other features. And as expected, conditional training does not encourage the BCDE to be aware of the distribution of $x$, resulting in a latent representation that is ignorant of the shift feature of $x$.

\section{Conclusion}
We presented a new framework for high-dimensional conditional density estimation. The building blocks of our framework are a pair of sibling models: the Bottleneck Conditional Density Estimator (BCDE) and the Bottleneck Joint Density Estimator (BJDE). These models use layers of stochastic neural networks as bottleneck between the input and output data. While the BCDE learns the conditional distribution $p(y \vert x)$, the BJDE learns the joint distribution $p(x, y)$. The bottleneck constraint implies that only the bottleneck needs to be marginalized when either the input $x$ or the output $y$ are missing during training, thus, enabling the BJDE to be trained in a semi-supervised fashion.

The key component of our framework is our hybrid objective function that regularizes the BCDE towards the BJDE. Our new objective is a novel extension of \citet{bishop:blend} that enables the principle of hybrid blending to be applied to deep variational models. Our framework provides a new mechanism for the BCDE, a conditional model, to become more robust and to learn from unlabeled data in semi-supervised conditional density estimation. 

Our experiments showed that hybrid training is competitive in the fully-supervised regime against pre-training, and achieves superior performance in the semi-supervised quadrant prediction task in comparison to conditional models, achieving new state-of-the-art performances on MNIST, SVHN, and CelebA. Even with pre-trained weight initializations, the conditional model is still susceptible to overfitting. In contrast, hybrid training is significantly more robust against overfitting. Furthermore, hybrid training transfers the nice embedding properties of the BJDE to the BCDE, allowing the BCDE to learn better and more robust representation of the input $x$. The success of our hybrid training framework makes it a prime candidate for other high-dimensional conditional density estimation problems, especially in semi-supervised settings. 
\clearpage
\bibliography{main}
\bibliographystyle{icml2017}
\clearpage
\appendix
\section{Factored Inference}\label{sec:factored}
When training the BJDE in the semi-supervised regime, we introduce a factored inference procedure that reduce the number of parameters in the recognition model. 

In the semi-supervised regime, the 1-layer BJDE recognition model requires approximating three posteriors: $p(z \vert x, y)\propto p(z)p(x,y|z)$, $p(z \vert x)\propto p(z)p(x|z)$, and $p(z \vert y)\propto p(z)p(y|z)$. The standard approach would be to assign one recognition network for each approximate posterior. This approach, however, does not take advantage of the fact that these posteriors share the same likelihood functions, i.e., $p(x,y|z)=p(x|z)p(y|z)$.

Rather than learning the three approximate posteriors independently, we propose to learn the approximate likelihood functions $\hat{\ell}(z ; x)\approx p(x|z)$, $\hat{\ell}(z; y)\approx p(y|z)$ and let $\hat{\ell}(z;x,y)=\hat{\ell}(z;x)\hat{\ell}(z;y)$. Consequently, this factorization of the recognition model enables parameter sharing within the joint recognition model (which is beneficial for semi-supervised learning) and eliminates the need for constructing a neural network that takes both $x$ and $y$ as inputs. The latter property is especially useful when learning a joint model over multiple, heterogeneous data types (e.g. image, text, and audio).

In practice, we directly learn recognition networks for $q(z \vert x)$ and $\hat{\ell}(z; y)$ and perform factored inference as follows
\begin{align}
q(z \vert x, y) \propto q_\phip(z \vert x)\hat{\ell}_\phip(z ; y), ~ q(z \vert y) \propto p(z)\hat{\ell}_\phip(z ; y),\label{eq:prop}
\end{align}
where $\phip$ parameterizes the recognition networks. To ensure proper normalization in \cref{eq:prop}, it is sufficient for $\hat{\ell}$ to be bounded. If the prior $p(z)$ belongs to an exponential family with sufficient statistics $T(z)$, we can parameterize $\hat{\ell}_\phip(z;y) = \exp\paren{\langle T(z), \eta_\phip(y)\rangle}$, where $\eta_\phip(y)$ is a network such that $\eta_\phip(y)\in \{\eta \vert \{\langle T(z), \eta\rangle\ \ \forall z\} \text{ is upper bounded}\}$. Then the approximate posterior can be obtained by simple addition in the natural parameter space of the corresponding exponential family. When the prior and approximate likelihood are both Gaussians, this is exactly precision-weighted merging of the means and variances \citep{sonderby:lvae}. 

\section{Derivation of the Hybrid Objective}\label{sec:derivation}
We first provide the derivation of \cref{eq:hybrid}. We begin with the factorization proposed in \cref{eq:bishop-blend}, which we repeat here for self-containedness,
\begin{align}
&p(\X_l, \Y_l, \X_u, \thetap, \theta) = p(\thetap, \theta)\nonumber\\
&\phantom{\hspace{1cm}}p(\X_u \vert \thetap) p(\X_l \vert \thetap) p(\Y_l \vert \X_l, \theta).
\label{eq:repeat}
\end{align}
Since our model includes unpaired $y$, we modify \cref{eq:repeat} to include
\begin{align}
&p(\X_l, \Y_l, \X_u, \Y_u, \thetap, \theta) = p(\thetap, \theta)\nonumber\\
&\phantom{\hspace{1cm}}p(\X_u \vert \thetap)p(\Y_u \vert \thetap) p(\X_l \vert \thetap) p(\Y_l \vert \X_l, \theta).
\end{align}
To account for the variational parameters, we include them in the joint density as well,
\begin{align}
&p(\X_l, \Y_l, \X_u, \Y_u, \thetap, \phip, \theta, \phi) = p(\thetap, \phip, \theta, \phi)\nonumber\\
&\phantom{\hspace{1cm}}p(\X_u \vert \thetap, \phip)p(\Y_u \vert \thetap, \phip)\nonumber\\
&\phantom{\hspace{1cm}}p(\X_l \vert \thetap, \phip) p(\Y_l \vert \X_l, \theta, \phi)
\label{eq:final-joint}
\end{align}
By taking the log and replacing the necessary densities with their variational lower bound,
\begin{align}
&\ln p(\X_l, \Y_l, \X_u, \Y_u, \thetap, \phip, \theta, \phi) \ge \ln p(\thetap, \phip, \theta, \phi) ~+ \nonumber \\
&\phantom{\hspace{1cm}} \J_x(\thetap, \phip; \X_u) + \J_y(\thetap, \phip; \Y_u) ~+ \nonumber \\
&\phantom{\hspace{1cm}} \J_x(\thetap, \phip; \X_l) + \C(\theta, \phi; \X_l, \Y_l),
\label{eq:hyb}
\end{align}
we arrive at \cref{eq:hybrid}. We note, however, that a more general hybrid objective \cref{eq:final-hybrid} is achievable. To derive the general objective, we consider an alternative factorization of the joint density in \cref{eq:final-joint},
\begin{align}
&p(\X_l, \Y_l, \X_u, \Y_u, \thetap, \phip, \theta, \phi) = p(\thetap, \phip, \theta, \phi, ) \nonumber\\
&\phantom{\hspace{1cm}}\phantom{={}} p(\X_l, \Y_l, \X_u, \Y_u \vert \thetap, \phip, \theta, \phi).
\label{eq:alt}
\end{align}
We factorize the likelihood term such that $\X_u$ and $\Y_u$ are always explained by the joint parameters $\thetap, \phip$,
\begin{align}
&p(\X_l, \Y_l, \X_u, \Y_u \vert \thetap, \phip, \theta, \phi) = p(\X_u \vert \thetap, \phip)p(\Y_u \vert \thetap, \phip)\nonumber\\
&\phantom{\hspace{1cm}}p(\X_l, \Y_l \vert \thetap, \phip, \theta, \phi).
\end{align}
We then introduce an auxiliary variable $s = \set{0, 1}$,
\begin{align}
&p(\X_l, \Y_l \vert \thetap, \phip, \theta, \phi) \nonumber\\
&\phantom{\hspace{1cm}} = \sum_{s} p(s)p(\X_l, \Y_l \vert \thetap, \phip, \theta, \phi, s),
\end{align}
where
\begin{align}
p(\X_l, \Y_l \vert \thetap, \phip, \theta, \phi, s^0) &= p(\X_l, \Y_l \vert \thetap, \phip)\\
p(\X_l, \Y_l \vert \thetap, \phip, \theta, \phi, s^1) &= p(\X_l \vert \thetap, \phip)p(\Y_l \vert \X_l, \theta, \phi).
\end{align}
Using Jensen's inequality, we can lower bound $\ln p(\X_l, \Y_l \vert \thetap, \phip, \theta, \phi)$ with 
\begin{align}
&p(s^0) \ln p(\X_l, \Y_l \vert \thetap, \phip) +  p(s^1) \ln p(\X_l \vert \thetap, \phip)p(\Y_l \vert \X_l, \theta, \phi).
\end{align}
By taking the log of \cref{eq:alt}, replacing all remaining densities with their variational lower bound, and setting $p(s^0) = \alpha$,
\begin{align}
&\ln p(\X_l, \Y_l, \X_u, \Y_u, \thetap, \phip, \theta, \phi) \nonumber\\
&\phantom{\hspace{0.3cm}}\ge \mathcal{H}(\thetap, \phip, \theta, \phi; \X_l, \Y_l, \X_u, \Y_u) \\
&\phantom{\hspace{0.3cm}}=\ln p(\thetap, \phip, \theta, \phi) ~+ \nonumber \\
&\phantom{\hspace{1cm}} \J_x(\thetap, \phip; \X_u) + \J_y(\thetap, \phip; \Y_u) ~+ \nonumber \\
&\phantom{\hspace{1cm}} \alpha\cdot \J_{xy}(\thetap, \phip; \X_l, \Y_l) ~+ \nonumber \\
&\phantom{\hspace{1cm}} (1 - \alpha)\cdot\brac{\J_x(\thetap, \phip; \X_l) + \C(\theta, \phi; \X_l, \Y_l)},
\label{eq:fin-hyb}
\end{align}
we arrive at the general hybrid objective. Note that when $\alpha = 0$, \cref{eq:fin-hyb} reduces to \cref{eq:hyb}.

\section{Visualizations for CelebA and SVHN}\label{sec:visualization}
We show visualizations of the hybrid BCDE predictions for CelebA and SVHN on the top-down prediction task in the $n_l = 10000$ semi-supervised regime. For each data set, we visualize both the images sampled during reconstruction as well as prediction using an approximation of the MAP estimate by greedily sampling the mode of each conditional distribution in the generative path.
\begin{figure}[!h]
\centering
\subfigure[Hybrid Rec.]{\includegraphics[width=0.2\textwidth]{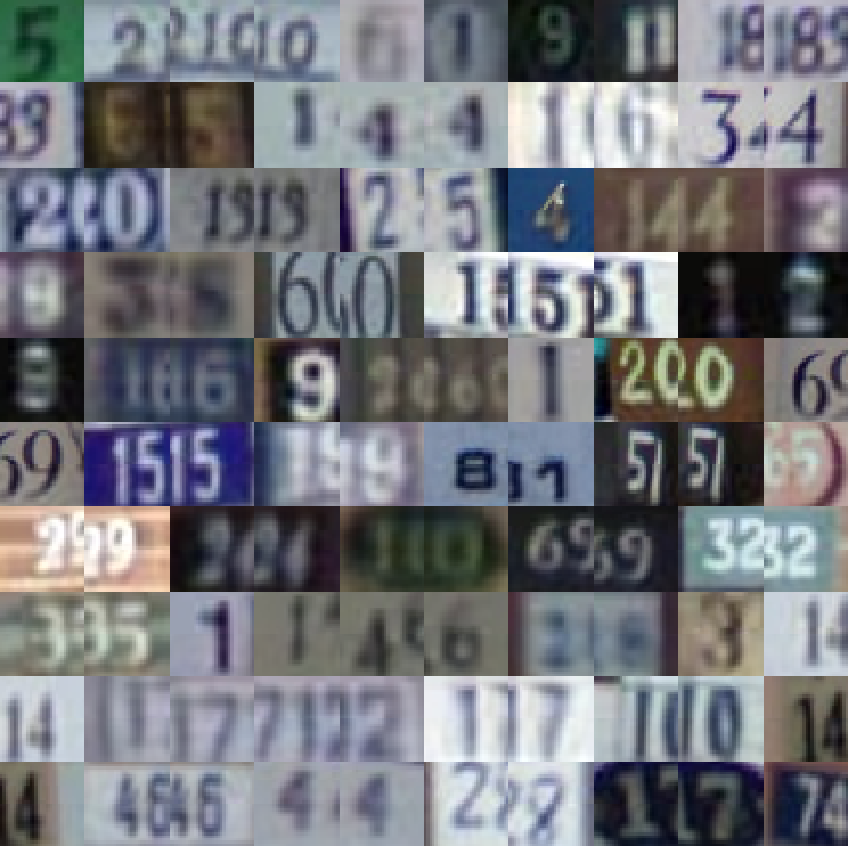}}
\subfigure[Hybrid Pred.]{\includegraphics[width=0.2\textwidth]{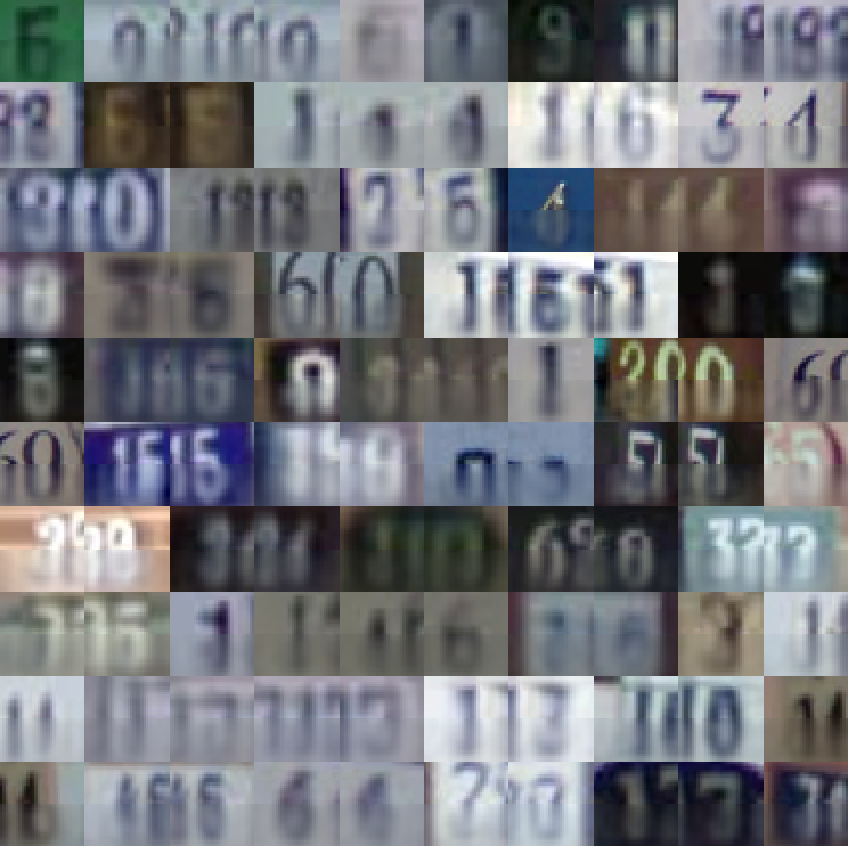}}
\caption{Visualization of the reconstructed and predicted bottom half of SVHN test set images when conditioned on the top half.}
\end{figure}

\begin{figure}[!t]
\centering
\subfigure[Conditional Rec.]{\includegraphics[width=0.2\textwidth]{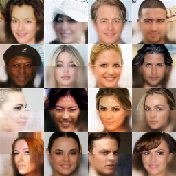}}
\subfigure[Conditional Pred.]{\includegraphics[width=0.2\textwidth]{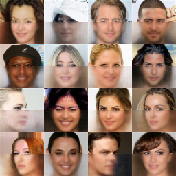}}
\subfigure[Pre-train Rec.]{\includegraphics[width=0.2\textwidth]{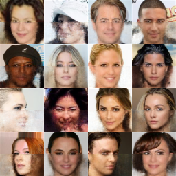}}
\subfigure[Pre-train Pred.]{\includegraphics[width=0.2\textwidth]{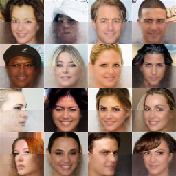}}
\subfigure[Hybrid Rec.]{\includegraphics[width=0.2\textwidth]{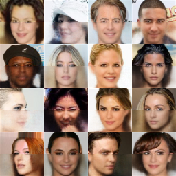}}
\subfigure[Hybrid Pred.]{\includegraphics[width=0.2\textwidth]{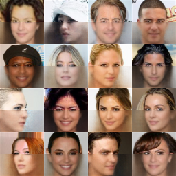}}
\subfigure[Hybrid Factored Rec.]{\includegraphics[width=0.2\textwidth]{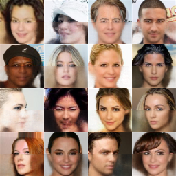}}
\subfigure[Hybrid Factored Pred.]{\includegraphics[width=0.2\textwidth]{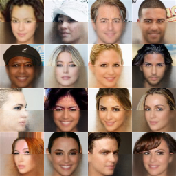}}
\caption{Visualization of the reconstructed and predicted bottom half of CelebA test set images when conditioned on the top half.}
\end{figure}
\end{document}